%% file: main.tex
\documentclass[letterpaper,10pt,conference]{ieeeconf}

\IEEEoverridecommandlockouts
\usepackage[utf8]{inputenc}
\DeclareUnicodeCharacter{0301}{\'{} }

\PassOptionsToPackage{dvipsnames}{xcolor}

\usepackage[backend=biber,style=ieee,url=false,doi=false]{biblatex}
\usepackage{problems}        
\usepackage{common}          
\usepackage{notation}        

\usepackage{tcolorbox}       
\usepackage[dvipsnames]{xcolor}  
\usepackage[font=footnotesize]{caption}  

\newcolumntype{Y}{>{\raggedright\arraybackslash}X}          
\newcolumntype{L}[1]{>{\raggedright\arraybackslash}p{#1}}   
\newcolumntype{M}{>{$}X<{$}}                                

\title{\LARGE \bf
Sparse Variable Projection in Robotic Perception:
Exploiting Separable Structure for Efficient Nonlinear Optimization
}

\def\authorInfo{
    Alan Papalia$^{1,2}$,
    Nikolas Sanderson$^{1}$,
    Haoyu Han$^{3}$,\\
    Heng Yang$^{3}$,
    Hanumant Singh$^{1}$,
    Michael Everett$^{1}$%
    \thanks{%
        Work was supported by the Northeastern University Institute for
        Experiential Robotics Postdoctoral Fellowship, Army Research Lab
        awards W911NF-24-2-006 and W911NF-24-2-0017,
        and Office of Naval Research grant N000142512322.
    }%
    \thanks{%
        $^{1}$Northeastern University, USA.
        {\tt\small \{sanderson.n, h.singh, m.everett\}@northeastern.edu}.
        $^{2}$University of Michigan.
        {\tt\small apapalia@umich.edu}.
        $^{3}$Harvard University.
        {\tt\small \{hyhan, hankyang\}@harvard.edu}.
    }%
}

\author{%
    \authorInfo
}

\begin{document}

\maketitle
\thispagestyle{empty}
\pagestyle{empty}

\input{sec/1_abstract.tex}
\input{sec/2_intro.tex}
\input{sec/3_related_work.tex}
\input{sec/4a_preliminary.tex}

\input{sec/5_experiments.tex}

\input{sec/6_conclusion.tex}

\renewcommand*{\bibfont}{\scriptsize}
\printbibliography

\end{document}

%% file: sec/1_abstract.tex
\begin{abstract}
Robotic perception often requires solving large nonlinear least-squares (NLS)
problems. While sparsity has been well-exploited to scale solvers, a
complementary and underexploited structure is \emph{separability} -- where some variables (e.g., visual
landmarks) appear linearly in the residuals and, for any estimate of the
remaining
variables (e.g., poses), have a closed-form solution.
Variable projection (VarPro) methods are a family of techniques that exploit
this structure by analytically eliminating the linear variables and presenting a
reduced problem in the remaining variables that has favorable properties.
However, VarPro has seen limited use in robotic perception; a major challenge
arises from gauge symmetries (e.g., cost invariance to global shifts and
rotations), which are common in perception and induce specific computational
challenges in standard VarPro approaches.
We present a VarPro scheme designed for problems with gauge symmetries that
jointly exploits separability and sparsity. Our method can be applied as a
one-time preprocessing step to construct a \emph{matrix-free Schur complement
operator}.
This operator allows efficient evaluation of costs, gradients, and
Hessian-vector products of the reduced problem and readily integrates with
standard iterative NLS solvers.
We provide precise conditions under which our method applies, and describe
extensions when these conditions are only partially met.
Across synthetic and real benchmarks in SLAM, SNL, and SfM, our approach
achieves up to \textbf{2$\times$--35$\times$ faster runtimes} than
state-of-the-art methods while maintaining accuracy. We release
an open-source C++ implementation and all datasets from our experiments.
\end{abstract}

%% file: sec/2_intro.tex
\section{Introduction}
\label{sec:introduction}

Robotic perception often requires solving large-scale optimization problems, typically
posed as nonlinear least-squares (NLS) problems with up to millions of variables
\cite{agarwal2010Bundle,ebadi2023present,kunze2018artificial,tranzatto2022cerberus}.
Such NLS formulations underpin key tasks such as simultaneous localization and
mapping (SLAM) \cite{cadena2017past}, structure from motion (SfM)
\cite{schonberger2016structure}, and sensor network localization (SNL)
\cite{mao2007wireless}.
In such tasks, a solver's performance directly impacts the
reliability and scale of robotic deployments
\cite{ebadi2023present,kunze2018artificial,tranzatto2022cerberus}.

While important, it is difficult to efficiently solve these problems because:
(i) they are large, often involving \(10^5\!-\!10^6\) variables \cite{agarwal2010Bundle};
(ii) applications typically require solving problems to high numerical accuracy
\cite{triggs2000Bundle}; and
(iii) problems often exhibit ill-conditioning that reduces the convergence rate of
standard methods \cite{golub2013matrix}.
To address these challenges, state-of-the-art systems
\cite{agarwal2022ceres,kummerle2011G2o,dellaert2012factor} exploit problem
sparsity to handle scale \cite{dellaert2017factor}, adopt second-order methods
(e.g., Levenberg--Marquardt) to ensure solution accuracy
\cite{nocedal2006numerical}, and employ numerically stable linear solvers to
mitigate ill-conditioning \cite{golub2013matrix}.
Despite these techniques, scaling limitations remain, motivating approaches to
further exploit problem structure.

\input{fig/title_fig.tex}

One path to improved efficiency is to exploit \emph{separable structure}
\cite{golub2003Separable}, in which least-squares residuals depend linearly on a
subset of unconstrained variables. Conditioning on the remaining variables
allows these linear variables to be eliminated in closed form, a process known
as variable projection (VarPro) \cite{golub2003Separable}. This reduction both
shrinks the problem size and improves convergence \cite{ruhe1980algorithms}.
Yet in many perception problems, gauge symmetries in the cost (e.g., globally
rotating the solution does not change the cost) introduce rank deficiencies in
the cost matrix. This rank deficiency led prior approaches
\cite{khosoussi2016Sparse} to adopt workarounds that limited efficiency gains.

\textbf{This paper.}
We show that a broad class of robotic perception problems (e.g.,
\cite{rosen2019SESync,papalia2024Certifiably,han2025Building,halsted22arxiv})
admits a VarPro scheme that simultaneously exploits sparsity and separability.
Our approach avoids challenges due to gauge symmetries and can be
applied as a one-time step before optimization begins.
We characterize this problem class by simple conditions on the cost and
constraints that can be checked a priori:
\begin{enumerate}[(i)]
      \item there must be a set of unconstrained variables -- these are the
            variables to be eliminated;
      \item the cost residuals must be linear functions; and
      \item the block of the cost residual Jacobian corresponding to the
            variables to be eliminated must have a specific graph-theoretic structure
            (see \cref{sec:prelim:efficient-schur-complement-products}).
\end{enumerate}
%
Our approach produces a matrix-free Schur complement
\cite{zhang2006schur}, i.e., an operator that can produce the reduced problem's
costs and gradients without forming the Schur complement (a large, dense
matrix).  This operator can be easily integrated into
standard iterative solvers, preserving the scalability of modern systems.
Finally, we discuss how the method extends when the ideal conditions are only
partially satisfied (e.g., when some residuals are nonlinear) and the
corresponding computational trade-offs.

Across a range of synthetic and real-world benchmarks in SLAM, SNL, and SfM, our
method consistently outperforms state-of-the-art baselines, achieving runtime
reductions of 2$\times$--35$\times$ and allowing for larger problems to be
solved within memory limits.
Our contributions are as follows:
\begin{enumerate}
      \item a novel sparsity-preserving VarPro scheme to accelerate a broad class
            of perception problems;
      \item a precise characterization of when this scheme applies, stated as
            simple conditions on costs and constraints; and
      \item an open-source C++ implementation and all datasets used in the experiments, for reproducibility.\footnote{Code: \href{https://github.com/UMich-RobotExploration/variable-projection}{github.com/UMich-RobotExploration}}
\end{enumerate}

%% file: fig/title_fig.tex
\begin{figure}[t]
    \centering
    \setlength{\fboxsep}{2.5pt}   
    \setlength{\fboxrule}{0pt}  

    \centering
    \includegraphics[width=0.99\linewidth]{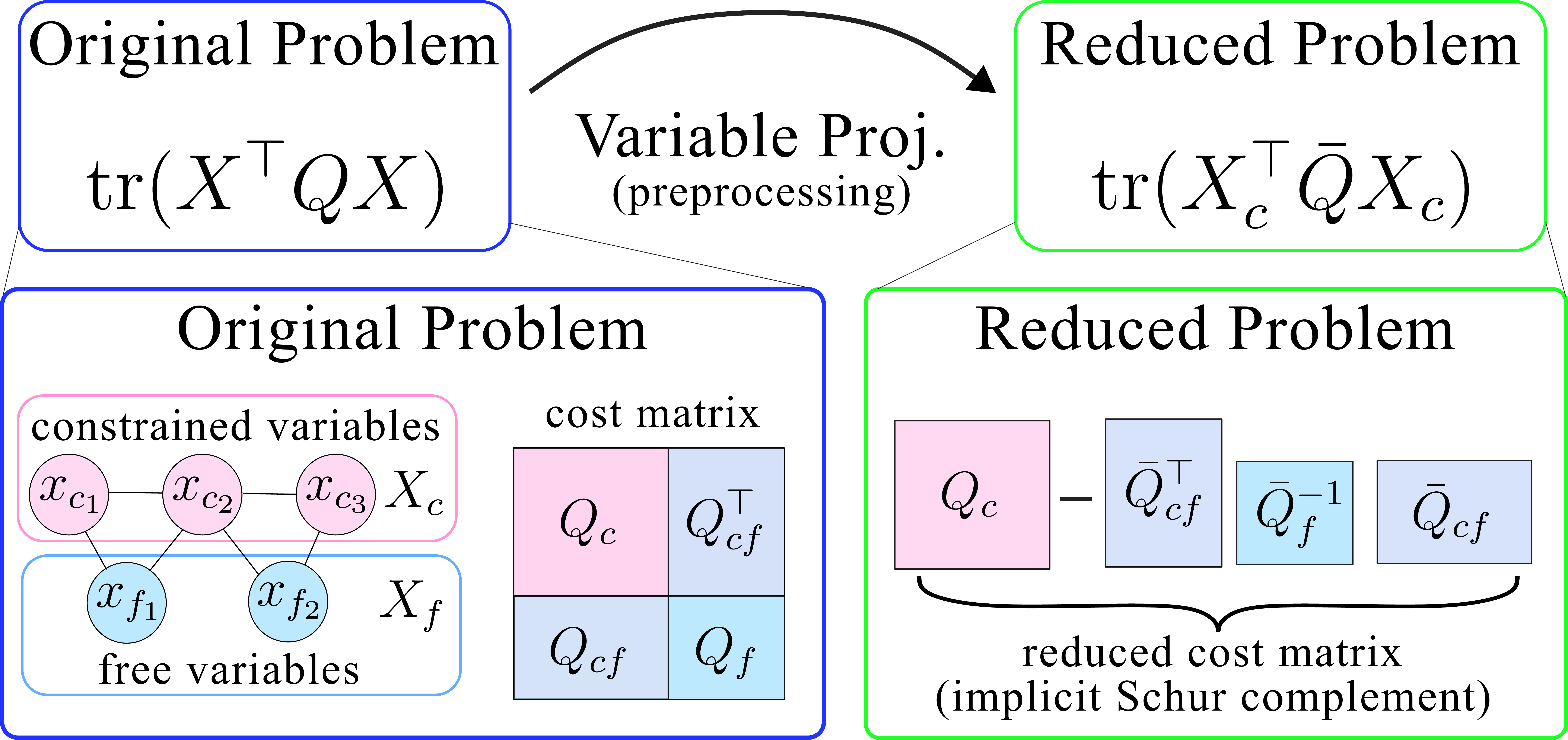}
    \fbox{}
    \vspace{-3mm}
    \hrule
    \fbox{}
    \includegraphics[width=0.99\linewidth]{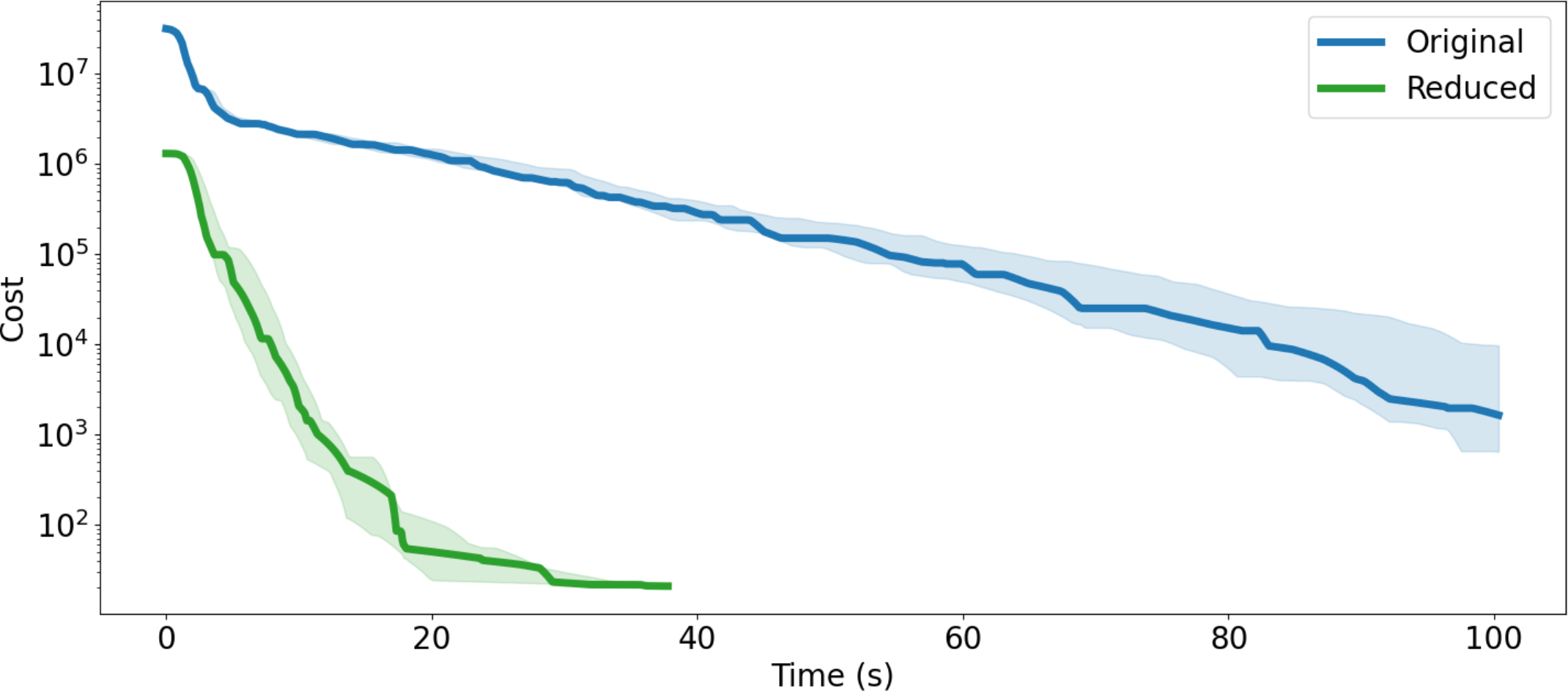}

    \caption{
        \textbf{(Top) method overview:}
        Our approach exploits \emph{separability} in optimization problem
        to perform variable projection and analytically eliminate a subset of
        variables (reducing problem size and improving conditioning),
        while preserving the efficiency of the original problem's sparsity
        structure.  Our approach can be applied as a one-time preprocessing step
        before passing the problem to a standard iterative solver.
        %
        \textbf{(Bottom) runtime improvement:}
        Comparison of cost vs.\ time for our reduced problem (green) and the
        original problem (blue) on a real-world structure from motion dataset
        (BAL-1934).  The variable projection step allows for substantial
        improvements in solver convergence and overall runtime.
    }
    \vspace{-5mm}
\end{figure}

%% file: sec/3_related_work.tex
\section{Related Work}
\label{sec:related-work}

This review focuses on works which use variable projection (VarPro) in robotic
perception.  We categorize methods by how the reduced problem is constructed.

We do not discuss other techniques to improve solver efficiency such
as general sparsity exploiting solvers
\cite{dellaert2012factor,kummerle2011G2o,agarwal2022ceres},
distributed computation
\cite{tian2020asynchronous,mcgann2024asynchronous,fan2023majorization}, or
initialization \cite{papalia2023score,carlone2015Initialization}.

For readers seeking broader background on VarPro, see
\textcite{golub2003Separable} for algorithms and applications and
\textcite{ruhe1980algorithms} for convergence analysis.

\textbf{Schur complement.}
The most common VarPro approach in practice is the \emph{Schur complement
trick}: eliminate the linear variables via block elimination of the normal
equations, leaving a reduced problem in the remaining variables
\cite{triggs2000Bundle,zhang2006schur}.
Schur methods come in two flavors.
{Explicit/direct Schur} methods fully form the reduced
matrix and optimize over the (dense) reduced problem.
This can be effective in SfM instances when the reduced problem is relatively
small \cite{han2025Building,woodford2020Large}.
%
%
{Implicit/matrix-free} methods do not form the reduced matrix; they apply it as
an operator via a series of matrix operations. This can better exploit problem
sparsity and has been used in SfM \cite{agarwal2010Bundle,weber2024Power} and SLAM
\cite{rosen2019SESync}.  For SLAM, \cite{khosoussi2016Sparse} gives a
sparsity-preserving scheme that does not use the Schur complement but is
iteration-equivalent to the Schur-based approach of \cite{barham1972algorithm}.

\textbf{Projection/QR.}
Instead of modifying normal equations, projection methods \emph{remove the effect
of the linear variables} by projecting the residuals onto a subspace that implies
the optimal assignment of the linear variables (obtained by thin QR
factorization). From this the residual only depends on the remaining variables
and optimization proceeds on that reduced problem.  This
has been used in appearance modeling \cite{matthews2004Active} and computer
vision applications of matrix factorization
\cite{hong2017Revisiting,okatani2011Efficient,hong2016Projective}.
While numerically stable, the projection step can be costly and it is more
difficult to exploit sparsity.

\textbf{Placement of our work.}
To our knowledge, only \cite{rosen2019SESync,khosoussi2016Sparse} consider
problems with gauge symmetries and try to preserve sparsity.
Our work generalizes \cite{rosen2019SESync} -- which focused on a specific
pose-graph optimization formulation -- through a different theoretical framework
that clarifies when the approach applies more broadly and how it can be extended
when conditions are only partially satisfied.
Our work is more limited in scope than \cite{khosoussi2016Sparse}, which
considers general separable residuals; our approach can be
viewed as gaining computational benefits over this method by restricting to a
stricter subset of residual functions.


%% file: sec/4a_preliminary.tex
\section{Problem Formulation}
\label{sec:problem-formulation}

Many robotic perception problems estimate variables (e.g., robot poses, landmark 
locations) from noisy measurements (e.g., odometry, visual observations, range 
measurements) by minimizing a sum of squared residuals, where each residual 
measures the discrepancy between a predicted and actual measurement.
We formalize this as a nonlinear least squares (NLS) problem, where the quantities
to estimate are represented as matrices.

\begin{problem}[Nonlinear Least Squares (NLS) Estimation]
\label{prob:nlls}
\begin{equation}
    \begin{alignedat}{2}
         & \min_{\matXc,\, \matXf} \quad &  & \quad \sum_{i=1}^{m} \|r_i(\matXc,\, \matXf) \|^2_{\matOmega_i} \\
         & \st                           &  & \quad \matXc \in \Xconstraint \subseteq \R^{n_c \times d}       \\
         &                               &  & \quad \matXf \in \R^{n_f \times d}
    \end{alignedat}
\end{equation}
where
$\matXc$ are variables constrained to the domain $\Xconstraint$,
$\matXf$ are unconstrained variables,
$r_i : \R^{n \times d} \to \R^{k_i \times d}$ is the residual function associated with the $i$-th measurement,
$\matOmega_i \in \PosDef^{k_i}$ is the positive definite concentration matrix
indicating the precision of the $i$-th measurement, and
$\|r_i(X)\|^2_{\matOmega_i} = \tr \left( r_i(X)^\top \matOmega_i r_i(X) \right)$ is a weighted squared Frobenius norm.
\end{problem}

This matrix-valued formulation generalizes the standard vector-valued NLS 
formulation, enabling us to address state-of-the-art perception problems 
involving constrained variables that are difficult to express in vector form 
(e.g., rotations satisfying $R^\top R = I$ 
\cite{rosen2019SESync,papalia2024Certifiably}).
We focus on a subclass of \cref{prob:nlls} where the residuals
are linear functions of the variables, enabling us to exploit
separable structure via a one-time preprocessing step.

\textbf{The case of linear residuals.}
\def\inhomogeneousFootnote{\footnote{We focus on the homogeneous
        case here for both simplicity and because there are no known inhomogeneous
        residuals that exhibit the graph structure we exploit in this work. However,
        the derivation for inhomogeneous residuals is straightforward, and we provide
        it in \cref{sec:appendix:quadratic-derivation}.}}
When residuals $r_i(X)$ are linear, the NLS problem
(\cref{prob:nlls}) reduces to a \emph{quadratic} cost.
We focus on \emph{homogeneous} linear residuals, i.e.,
$r_i(X) = A_i X$ for some matrix $A_i \in \R^{k_i \times n}$,
as (i) it simplifies notation, (ii)
many common perception residuals are homogeneous (see
\cref{tab:quadratic-factors}), and (iii) inhomogeneous residuals can be
made homogeneous through variable augmentation
\cite{luo2010semidefinite}.

By vertically stacking the Jacobians of the residuals $A_i$ and block-diagonally
arranging the concentration matrices $\matOmega_i$, we can rewrite the cost as
\begin{align}
     & \sum_i \big\| r_i(\matX)\big\|_{\matOmega_{i}}^2 \,=\, \,\big\|
    \matA\, \matX \big\|_{\matOmega}^2 \,=\, \tr \left(
    \matX^\top \matA^\top  \matOmega \matA
    \matX \right),                                                                             \\
     & \matA \triangleq \begin{bmatrix}\matA_1^\top \cdots \matA_m^\top \end{bmatrix}^\top ,\; \\
     & \matOmega \triangleq \text{blkdiag}(\matOmega_1, \ldots, \matOmega_m),                  \\
     & \matX \triangleq \begin{bmatrix}
                            \matXc \\
                            \matXf
                        \end{bmatrix} \in \begin{bmatrix}
                                              \Xconstraint \subseteq \R^{d \times n_c} \\
                                              \R^{d \times n_f}
                                          \end{bmatrix}, \quad n = n_c + n_f,
\end{align}
where the $\text{blkdiag}(\cdot)$ operator constructs a block-diagonal matrix
from its arguments.

\input{tab/common_residuals.tex}

We can reformulate the NLS problem (\cref{prob:nlls}) as
follows, where $\matQ \triangleq \matA^\top \matOmega \matA$:
\begin{problem}[Constrained Quadratic Cost Problem]
\label{prob:general-qcqp}
\begin{equation}
    \begin{alignedat}{2}
         & \min_{X \in \R^{d \times n}} & \tr(X^\top Q X)          \\
         & \st                          & \matXc \in \Xconstraint,
    \end{alignedat}
\end{equation}
\end{problem}

Importantly, \cref{prob:general-qcqp} possesses a globally quadratic cost (i.e.,
is quadratic without linearization). This enables VarPro as a one-time preprocessing
step, as shown next.

\section{Separable Structure \& Variable Projection}
\label{sec:prelim:separable-structure}

Separable structure in an optimization problem refers to
the situation where, if certain variables are held fixed, the remaining
variables can be efficiently solved in closed-form. This structure partitions
the variables into two sets: those that are more difficult to optimize over
(e.g., due to constraints) and those that are easily optimized.

Variable projection (VarPro) methods
\cite{golub2003Separable} exploit separable structure by
iteratively optimizing over the `difficult' variables while implicitly
considering the `easy' variables at their optimal values conditioned on the
current iterate of the `difficult' variables.
VarPro effectively reduces the dimensionality of the problem (to just the
`difficult' variables), and has been both theoretically and empirically shown to
improve convergence rates of optimization
\cite{ruhe1980algorithms}.

The problems we consider exhibit separable
structure that is particularly well suited for VarPro.  Because
the `easy' variables are unconstrained and the cost is quadratic, a
closed-form elimination of the unconstrained variables can be performed once and
holds at every iteration.
We now show how unconstrained variables in a quadratic cost induce separable
structure and how the Schur complement can eliminate these
variables.

With the variable ordering $X = [\matXc^\top \mid \matXf^\top]^\top$, we can
partition the matrix $Q$ as
\begin{equation}
    \label{eq:Q-partition}
    \matQ = \begin{bmatrix}
        \matQcc & \matQcf \\
        \matQfc & \matQff
    \end{bmatrix}.
\end{equation}
For a fixed $\matXc$, we set the gradient of the cost
with respect to the unconstrained variables $\matXf$ to zero and solve for the
optimal value $\matXfstar$ as a function of $\matXc$:
\begin{align}
     & \frac{\partial}{\partial \matXf} \tr(X^\top Q X) = 2 \matQff \matXf + 2 \matQfc \matXc = 0, \\
     & \matXfstar = - \matQff^{\dagger} \matQfc \matXc,
\end{align}
where $\matQff^{\dagger}$ is the Moore-Penrose pseudoinverse of $\matQff$.

Plugging this optimal value $\matXfstar$ into the cost, we obtain a cost
that depends only on the constrained variables $\matXc$:
\begin{equation}
    \label{eq:reduced-cost}
    \tr(\matXc^\top \matQmarg \matXc)  =
    \tr\left(\matXc^\top \left(\matQcc - \matQcf \matQff^{\dagger} \matQfc\right) \matXc\right),
\end{equation}
where $\matQmarg \triangleq \matQcc - \matQcf \matQff^{\dagger} \matQfc$
is commonly called the \emph{Schur complement} \cite{zhang2006schur} of
$\matQff$ in $\matQ$.

This leads to the following reduced problem, which depends only on the constrained
variables $\matXc$,
\begin{problem}[\emph{Reduced} Constrained Quadratic Cost Problem]
\label{prob:reduced-qcqp}
\begin{equation}
    \min_{\matXc \in \Xconstraint} \quad \tr(\matXc^\top \matQmarg \matXc)
\end{equation}
\end{problem}

\textbf{Computational challenges in the reduced cost.}
%
The steps above yield a reduced problem (\cref{prob:reduced-qcqp}) in
the constrained variables $\matXc$.  However, two obstacles appear:
(i) forming the Schur complement $\matQmarg$ requires a pseudoinverse, which is
costly and numerically fragile; and (ii) $\matQmarg$ is typically dense even
when the original matrix $\matQ$ is sparse, making storage and operations
expensive. Either issue can erase the benefits of eliminating the unconstrained
variables. We avoid these challenges by leveraging iterative methods
that do not require explicitly forming $\matQmarg$.

\textbf{Proposed approach: implicit (matrix-free) Schur via iterative methods.}
Rather than explicitly forming the dense Schur complement $\matQmarg$,
we solve the reduced problem (\cref{prob:reduced-qcqp}) using iterative
methods \cite{saad2003iterative}, which only require computing
matrix-vector products $\matQmarg\,\matXc$.
Our key contribution is an efficient, matrix-free routine for computing
these products without ever forming $\matQmarg$ explicitly or computing
a pseudoinverse.
Specifically, we derive an exactly equivalent reformulation of $\matQmarg$
that can be applied via sparse matrix operations and triangular solves
with Cholesky factors of the original problem data.

\section{Matrix-Free Schur Complement Products}
\label{sec:prelim:focus-schur-complement-products}

\def\matAf{\matA_{\text{f}}}
\def\matAc{\matA_{\text{c}}}
\def\matAfT{\matA_{\text{f}}^{\top}}
\def\matAcT{\matA_{\text{c}}^{\top}}

This section identifies the specific term that creates computational
challenges in computing Schur complement products $\matQmarg \matXc$, namely a
pseudoinverse that creates a large, dense matrix. We then show how
least-squares structure admits a reformulation that
replaces the pseudoinverse with a positive definite matrix inverse, which can be
efficiently represented via Cholesky factorizations. This reformulation
allows us to compute products with $\matQmarg$ via a series of sparse matrix
products and triangular solves.

We revisit the quadratic cost matrix $\matQ = \matA^\top {\Omega} \matA$.  By
ordering the variables as $X = [\matXc^\top \mid \matXf^\top]^\top$, the
Jacobian matrix becomes $\matA = [\matAc \mid \matAf ]$, where $\matAc$ and
$\matAf$ are the stacked Jacobians of the residuals with respect to the
constrained and unconstrained variables, respectively. This then reframes
$\matQ$ and the Schur complement $\matQmarg$ as:
\begin{align}
     & \matQ = \begin{bmatrix}
                   \matQcc                  & \matAcT \matOmega \matAf \\
                   \matAfT \matOmega \matAc & \matAfT \matOmega \matAf
               \end{bmatrix}, \\
     & \matQmarg = \matQcc -
    \matAcT \matOmega \matAf
    (\matAfT \matOmega \matAf)^{\dagger}
    \matAfT \matOmega \matAc.
\end{align}

\def\matQtwo{\matQ_{2}}
Importantly, $\matQ$ and the constituent matrices
$\matQcc,\matAc,\matAf,\matOmega$ naturally inherit the sparsity of the
graphical structure of the underlying problem.
%
We can write products with the Schur complement $\matQmarg \matXc$ as:
\begin{align}
    \label{eq:schur-product-dense-pseudoinverse}
     & \matQmarg \matXc =
    \underbrace{\matQcc}_{\text{sparse}} \matXc
    -
    \underbrace{
        \Big(
        \underbrace{\matAcT \matOmega \matAf}_{\text{sparse}}
        \underbrace{
            (\matAfT \matOmega \matAf)^{\dagger}
        }_{\text{dense}}
        \underbrace{\matAfT \matOmega \matAc}_{\text{sparse}}
        \Big)
    }_{\matQtwo}
    \matXc,                \\
     & \matQtwo \triangleq
    \matAcT \matOmega \matAf
    (\matAfT \matOmega \matAf)^{\dagger}
    \matAfT \matOmega \matAc,
\end{align}
where we have highlighted the sparsity of each term
assuming the original problem is sparse.

\textbf{Key challenge: dense pseudoinverse in Schur complement products.}
\cref{eq:schur-product-dense-pseudoinverse} emphasizes that the pseudoinverse
$(\matAfT \matOmega \matAf)^{\dagger}$ creates a dense matrix that is the
computational bottleneck in computing products with the Schur complement.
If $(\matAfT \matOmega \matAf)$ were full rank and positive definite,
as in many VarPro applications \cite{golub2003Separable},
the pseudoinverse becomes an inverse. Since $(\matAfT \matOmega
    \matAf)$ inherits the sparsity of the residuals, this inverse could be
efficiently computed via sparse Cholesky factorization and
applied via sparse triangular solves. However, in many robotic perception
problems the matrix is rank-deficient due to inherent symmetries.

\subsection{Exact Reformulation of $\matQtwo$ via CR Decomposition}
\label{sec:prelim:graphical-structure}


Our reformulation relies on the $CR$ decomposition of a matrix
\cite{strang2022Three,strang2024Elimination}, which factorizes a matrix by its
column and row spaces and can be computed e.g., via rank-revealing QR
decomposition \cite{golub2013matrix}.
We first factorize $\matAf$ as
\begin{equation}
    \matAf = \matC \matR,
\end{equation}
where the columns of $\matC$ form a basis for the column space of $\matAf$ and
the rows of $\matR$ form a basis for the row space of $\matAf$.  The matrix
$\matC$ has full column rank and the matrix $\matR$ has full row rank.
%
Using this decomposition, we can rewrite the inner portion of $\matQtwo$ as
\begin{equation}
    \label{eq:Q2-reformulation-start}
    \begin{aligned}
        \matAf (\matAfT \matOmega \matAf)^{\dagger} \matAfT
        =  (\matC \matR) (\matR^\top \matC^\top \matOmega \matC \matR)^{\dagger} (\matC \matR)^\top
    \end{aligned}
\end{equation}
Since $\matC$, $\matR$, and $\matOmega$ are all full rank, we use the
pseudoinverse property $(A \matOmega B)\pinv = B\pinv \matOmega\pinv A\pinv$ to
rewrite \cref{eq:Q2-reformulation-start} as
\begin{equation}
    \label{eq:Q2-reformulation-middle}
    =
    \matC
    (\matR \matR\pinv)
    (\matC^\top \matOmega \matC)^{\dagger}
    ((\matR^\top)\pinv \matR^\top)
    \matC^\top.
\end{equation}
Due to the full column rank of $\matR$, we can apply the properties
$(\matR \matR\pinv) = I$ and $(\matR^\top)\pinv \matR^\top = I$ to simplify
\cref{eq:Q2-reformulation-middle} to
\begin{equation}
    =
    \matC (\matC^\top \matOmega \matC)^{\dagger} \matC^\top.
\end{equation}

Since $\matC$ has full column rank and $\matOmega$ is positive definite, the
matrix $\matC^\top \matOmega \matC$ is positive definite. As a result, we can
replace the pseudoinverse with the inverse and represent
the inverse via Cholesky factorization ($\matC^\top \matOmega \matC = LL^\top$)
\begin{align}
     & = \matC (\matC^\top \matOmega \matC)^{-1} \matC^\top    \\
     & = \matC L^{-\top} L^{-1} \matC^\top \label{eq:Q2-final}
\end{align}
Plugging this reformulation into $\matQtwo$ in
\cref{eq:schur-product-dense-pseudoinverse} obtains
\begin{equation}
    \label{eq:Qmarg-product-final}
    \begin{aligned}
        \matQmarg \matXc
         & =
        \matQcc \matXc -
        \left( \matAcT \matOmega \matC \right)
        L^{-\top} L^{-1}
        \left(\matC^\top \matOmega \matAc\right)
        \matXc \\
         & =
        \matQcc \matXc - \left(B L^{-\top} L^{-1} B^\top\right) \matXc,
    \end{aligned}
\end{equation}
where $B \triangleq \matAcT \matOmega \matC$.
This can be performed as a series of matrix products and forward- and
back-substitution with the Cholesky factors $L$ and $L^{\top}$ as described in
\cref{alg:schur-complement-product}.

\textbf{When is this reformulation valid?}
The reformulation in \cref{eq:Qmarg-product-final} is exact and only
depends on the least-squares structure (which induces the block
structuring of $\matQ$ in \cref{eq:Q-partition}). This could be applied at each
iteration of an iterative solver that linearizes the problem,
though this incurs the cost of recomputing $\matC$ and the Cholesky
factorization. Because our problems have linear
residuals, we perform this reformulation once as preprocessing.

\textbf{When is the Schur product in \cref{eq:Qmarg-product-final} efficient?}
Efficiency is driven by the sparsity of
$\matC$, which determines the density of both $B=\matAcT \matOmega \matC$
and the Cholesky factors $L^{-\top} L^{-1}$.
As $\matC$ is a linearly independent basis for the column space of $\matAf$,
there are many ways to compute it with sparsity in mind \cite{coleman1987null}.
These approaches typically require either non-trivial
computation (e.g., factorization) or \emph{a priori} knowledge of the rank of
$\matAf$ (e.g., to determine stopping in greedy algorithms).

\input{alg/schur_complement_product.tex}

\subsection{Leveraging Graph Structure for Efficient Schur Products}
\label{sec:prelim:efficient-schur-complement-products}

We now establish specific, yet common, graph-theoretic conditions on
the Jacobian $\matAf$ under which the implicit Schur complement product in
\cref{eq:Qmarg-product-final} retains the sparsity of the original problem.
Furthermore, these conditions allow for closed-form computation of $C$
and $(\matC^\top \matOmega \matC)$ by simply removing rows and columns
of the original matrices $\matAf$ and $(\matAfT \matOmega \matAf)$,
making the reformulation in \cref{eq:Qmarg-product-final} efficient.

Consider a graph $G$ where each unconstrained variable is a node and any
two variables that appear together in a residual are connected by an edge.
Many robotic perception problems yield a connected graph over the unconstrained
variables.

Furthermore, for
residuals based on relative differences between pairs of unconstrained
variables (i.e., $r_i(\matX) = g(\matXc) (\matX_{fi} - \matX_{fj})$ for some
function $g(\matXc)$), the Jacobian $\matAf$ can be interpreted as the
\emph{incidence matrix} of this graph $G$.  I.e., each row of $\matAf$
corresponds to an edge in $G$ and each column to a
variable (node). For a row corresponding to an edge
between nodes $i$ and $j$, the $k$-th entry is
\begin{equation}
    \matA_{\text{f}, (i,j)} (k) = \begin{cases}
        1,  & k = i            \\
        -1, & k = j            \\
        0,  & \text{otherwise}
    \end{cases}.
\end{equation}

\def\rankDeficientFootnote{\footnote{The rank-deficiency of $\matAf$ is because
        the graphs inherently capture relative information. This can be seen
        from the fact that the all-ones vector is the null space of $\matAf$;
        adding a constant offset to all variables does not affect relative
        differences.}}

We leverage one key property of incidence matrices for connected graphs
\cite{chung1997spectral}:
removing any one column of the incidence matrix yields a linearly independent
basis for the column space of the matrix.\rankDeficientFootnote

Furthermore, the matrix $\matAfT \matOmega \matAf$ is the \emph{weighted graph
    Laplacian} of the graph, which is positive semidefinite with a
single zero eigenvalue \cite{chung1997spectral}. Closely related to the property
above, the product $\matC^\top \matOmega \matC$ when $\matC$ is a reduced
incidence matrix is called a \emph{reduced graph Laplacian} and is positive
definite \cite{chung1997spectral}. The reduced graph Laplacian is
formed by removing the same row and column from the graph Laplacian
that was removed from $\matAf$ to form $\matC$.

As both the reduced incidence matrix and reduced graph Laplacian are formed by
removing rows and columns of the original matrices $\matAf$ and $(\matAfT
\matOmega \matAf)$, they are easily constructed and inherit the sparsity of the
original matrices. As a result, the reduced incidence matrix is a natural choice
for $\matC$ in \cref{eq:Qmarg-product-final}. With this choice, the only
non-negligible computation required to form the operator for the Schur
complement product (as in \cref{alg:schur-complement-product}) is a (sparse)
Cholesky factorization of the reduced graph Laplacian.

In \cref{fig:schur-matrix-free}, we visualize the sparsity patterns of the
matrices involved in the Schur complement product $\matQmarg \matXc$
for a specific problem instance. In contrast to forming
the dense Schur complement $\matQmarg$, the reformulation in
\cref{eq:Qmarg-product-final} retains the sparsity of the original problem.

If $A_f$ is not incidence-like (i.e., $\matXf$ does not appear in residuals
solely through pairwise differences of its entries), $\matC$ can instead be
computed using basis-construction techniques \cite{coleman1987null}. This may
incur additional computational cost and yield a denser $\matC$, but the
reformulation in \cref{eq:Qmarg-product-final} remains exact.

\input{tab/dense_vs_sparse_marg.tex}

%% file: tab/common_residuals.tex
\renewcommand{\arraystretch}{1.5}
\begin{table}[t]
    \vspace{0.5em}
    \caption{Example linear residuals in state-of-the-art formulations.
        In these problems the variables are:
        $R_i$ (rotation matrices),
        $t_i$ (translation vectors),
        $u_{ij}$ (unit bearing vectors), and
        $RS_{ij}$ (scaled rotation matrices).
        The quantities
        $\tilde{R}_{ij}, \tilde{t}_{ij}, \tilde{u}_{ij}$ are noisy measurements of the same
        objects and $\tilde{d}_{ij}$ are noisy distance measurements.
        All of these residuals are homogeneous linear functions (i.e., $r_{ij}(\matX) = A_{ij} \matX$).
    }
    \label{tab:quadratic-factors}
    \centering
    \begin{tabularx}{\linewidth}{XX}
        \toprule
        \textbf{Measurement Type}                              & \textbf{Residual $r_{ij}(\matX)$}     \\
        \midrule
        Relative rotation   \cite{rosen2019SESync}             & $R_j - R_i\,\tilde{R}_{ij}$           \\
        Relative translation \cite{rosen2019SESync}            & $t_j - t_i - R_i\,\tilde{t}_{ij}$     \\
        Scale-free relative translation \cite{han2025Building} & $t_j - t_i - RS_i\,\tilde{t}_{ij}$    \\
        Range  \cite{papalia2024Certifiably,halsted22arxiv}    & $t_j - t_i - u_{ij}\, \tilde{d}_{ij}$ \\
        \bottomrule
    \end{tabularx}
    \vspace{-1.5em}
\end{table}

%% file: alg/schur_complement_product.tex
\begin{algorithm}[t]
    \caption{Matrix-Free Schur Complement Products}
    \label{alg:schur-complement-product}
    \textbf{Input:} Matrices $\matQcc$, $\matAc$, $\matAf$, $\matOmega$, and $\matXc$. \\
    \textbf{Output:} Product $\matQmarg \matXc$

    \textbf{Precomputation} (once, given $\matQcc$, $\matAc$, $\matAf$, $\matOmega$):
    \begin{algorithmic}[1]
        \State $\matC \gets \text{CR} (\matAf)$. \Comment{CR decomposition (\cref{sec:prelim:focus-schur-complement-products})}
        \State $L \gets \text{Cholesky}(\matC^\top \matOmega \matC)$. \Comment{sparse factorization}
        \State $B \gets \matC^\top \matOmega \matAc$.
    \end{algorithmic}
    \textbf{Online computation} (per new $\matXc$):
    \begin{algorithmic}[1]
        \State $Y \gets B \matXc$ \Comment{sparse matrix product}
        \State $Z \gets L^{-1} (L^{-\top} Y)$ \Comment{sparse triangular solves}
        \State \Return $\matQmarg \matXc = \matQcc \matXc - B^\top Z$.
    \end{algorithmic}
\end{algorithm}

%% file: tab/dense_vs_sparse_marg.tex
\setlength{\fboxsep}{0pt}%
\setlength{\fboxrule}{0.5pt}%
\newlength{\cholsize}
\setlength{\cholsize}{0.35cm}

\newcommand{\MatrixImg}[2]{%
    \fbox{\adjustbox{valign=c}{%
            \includegraphics[height=\dimexpr #1\cholsize\relax]{#2}%
        }}%
}

\newcommand{\QMainSparse}{\MatrixImg{4}{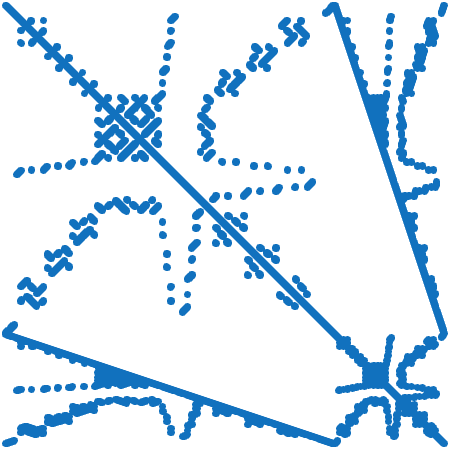}}
\newcommand{\QOneSparse}{\MatrixImg{3}{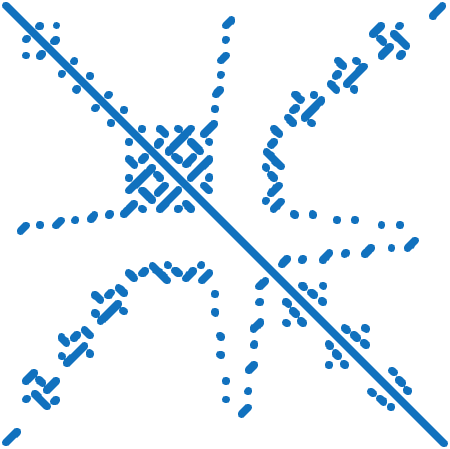}}
\newcommand{\QTwoDense}{\MatrixImg{3}{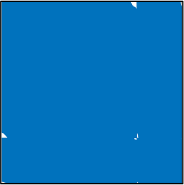}}
\newcommand{\BSparse}{\MatrixImg{3}{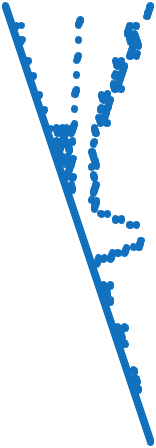}}
\newcommand{\LSparse}{\MatrixImg{1}{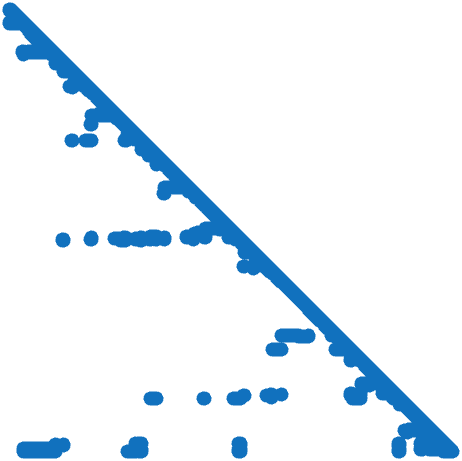}}

\def\SparseOp{\left( \BSparse \underbrace{ \left(\LSparse\right)^{\shortminus 1} \left(\LSparse\right)^{\shortminus\top} }_{\text{Cholesky Factors}} \BSparse^\top \right)}
\newcommand{\wrapWords}[2]{\makecell[l]{#1 \\ #2}}

\def\SparseFullOp{\QOneSparse  - \SparseOp }
\def\DenseFullOp{\QTwoDense  }

\renewcommand{\arraystretch}{1}
\begin{figure}[t]
    \vspace{0.5em}
    \centering
    \begin{tabularx}{\linewidth}{L{1.7 cm} M}
                                                                          & \textbf{Schur Complement Products} \\
        \midrule
        \wrapWords{\textbf{Original}}{\textbf{Matrix} $\mathbf{(\matQ)}$} & \QMainSparse                               \\
        \midrule
        \wrapWords{\textbf{Explicit}}{\small{(Dense)}}                    & \DenseFullOp                               \\ \addlinespace[4pt]
        \wrapWords{\textbf{Implicit}}{\small{(Sparse)}}                   & \SparseFullOp                              \\
        \bottomrule
    \end{tabularx}
    \caption{
        \textbf{Matrix-Free Schur Complement Products:}
        Here we demonstrate the difference between explicitly forming the Schur
        complement $\matQmarg$ as a dense matrix versus performing a series of
        sparse operations to implicitly compute the product $\matQmarg \matXc$
        without forming the matrix.
        The top row shows the sparsity pattern of the original matrix $\matQ$.
        The bottom two rows show the sparsity patterns of the dense (explicit)
        and sparse (implicit) Schur complement approaches with the data
        from the `Garage' dataset \cite{carlone2015Initialization}.
    }
    \label{fig:schur-matrix-free}
    \vspace{-1.5em}
\end{figure}

%% file: sec/5_experiments.tex
\section{Experiments}
\label{sec:cora-marg:computational-experiments}

\input{tab/tab_experiments.tex}

We analyze how our methodology impacts computational efficiency for four common 
state estimation problems: pose-graph optimization (PGO), range-aided SLAM 
(RA-SLAM), structure from motion (SfM), and sensor network localization (SNL). 
We consider both \emph{iteration count} and \emph{wall-clock time} to convergence.

All experiments use residuals from \cref{tab:quadratic-factors}, where our 
methodology fully exploits the problem structure: cost residuals are homogeneous 
linear functions (\cref{sec:problem-formulation}) and the Jacobian block $\matAf$ 
corresponding to unconstrained variables is a directed incidence matrix
(\cref{sec:prelim:efficient-schur-complement-products}).

\subsection{Implementation and Baselines}

\textbf{Our approach.}
We implement our approach within a Riemannian trust-region (RTR)
framework~\cite{absil2007trust} using preconditioned truncated conjugate
gradients (pTCG)~\cite[Ch. 6.5]{boumal2023introduction} for trust-region 
subproblems. We use the matrix-free Schur complement product algorithm 
(\cref{sec:prelim:efficient-schur-complement-products}) to efficiently 
compute matrix-vector products.
We precondition pTCG with the \emph{regularized Cholesky preconditioner}
\cite[Sec. VI.A]{papalia2024Certifiably} of $\matQ$: $P = L^{\shortminus \top} 
L^{\shortminus 1} = (\matQ + \mu I)^{\shortminus 1}$, where $\mu$ is chosen so 
the condition number of $P$ is below $10^6$. The preconditioner is stored as a 
sparse Cholesky factor $L L^{\top} = (\matQ + \mu I)$ computed once at 
initialization. Since the reduced variable $\matXc$ does not match the 
preconditioner dimensions, we bottom-pad with zeros ($\matX = [\matXc; 0]$).

\textbf{Baselines.}
We benchmark against three baselines: (i) \emph{Original}, which directly 
optimizes the full problem without variable elimination; (ii) \emph{Original + 
VarPro}~\cite{khosoussi2016Sparse}, which optimizes the same problem as 
\emph{Original} but uses variable projection to update unconstrained variables 
in closed form at each iteration; and (iii) the Levenberg-Marquardt 
implementation in GTSAM~\cite{dellaert2012factor}, a state-of-the-art solver 
using direct factorization-based linear solvers.

The first two baselines (Original and Original + VarPro) are implemented as
alternative options within the same RTR framework as our approach. Our GTSAM
implementations uses custom manifolds and residuals from \cite{certi_fgo} to
ensure identical problem formulations.

\textbf{Hardware and software.}
Experiments were conducted on a laptop running Ubuntu 20.04 with an Intel
i7-12700H CPU (20 threads) and 40\,GB of RAM. All implementations are available as
open-source C++ libraries along with data and scripts to reproduce the
experiments.
\begin{figure*}[t]
    \vspace{0.5em}
    \centering
    \subfloat[PGO (Intel)\label{fig:a}]{
        \includegraphics[width=0.48\linewidth]{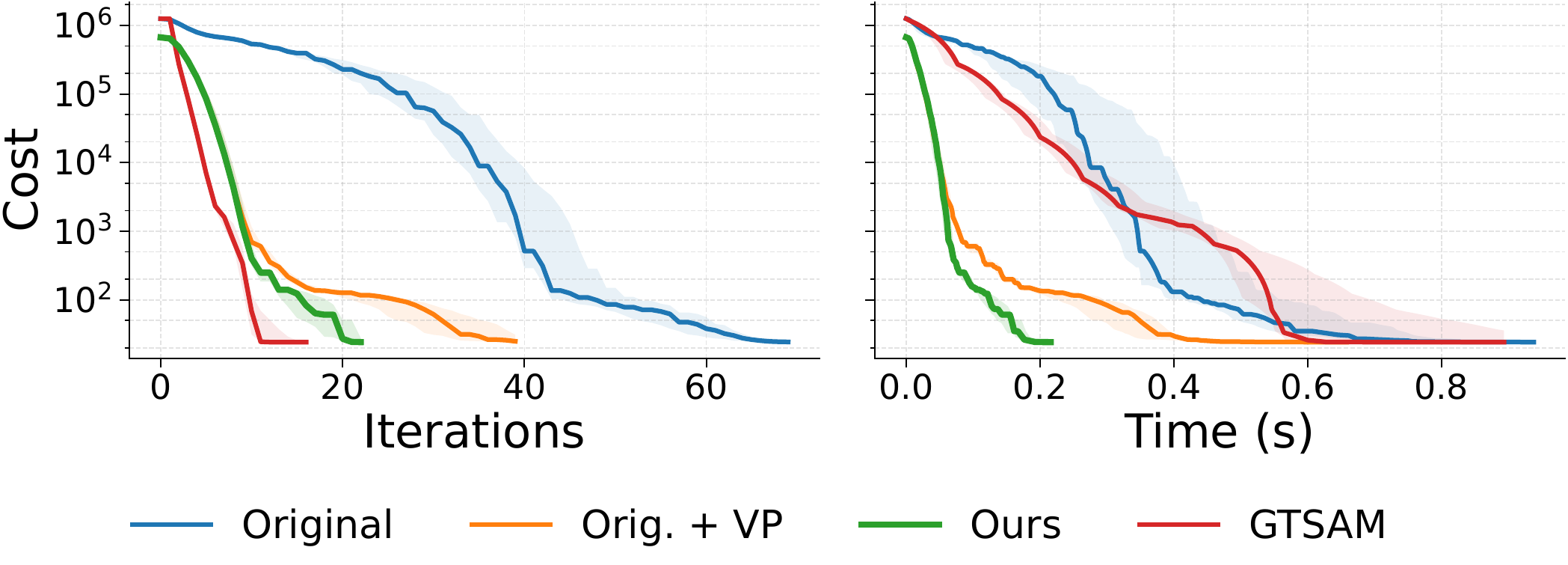}
    }\hfill
    \subfloat[RA-SLAM (Single Drone)\label{fig:b}]{
        \includegraphics[width=0.48\linewidth]{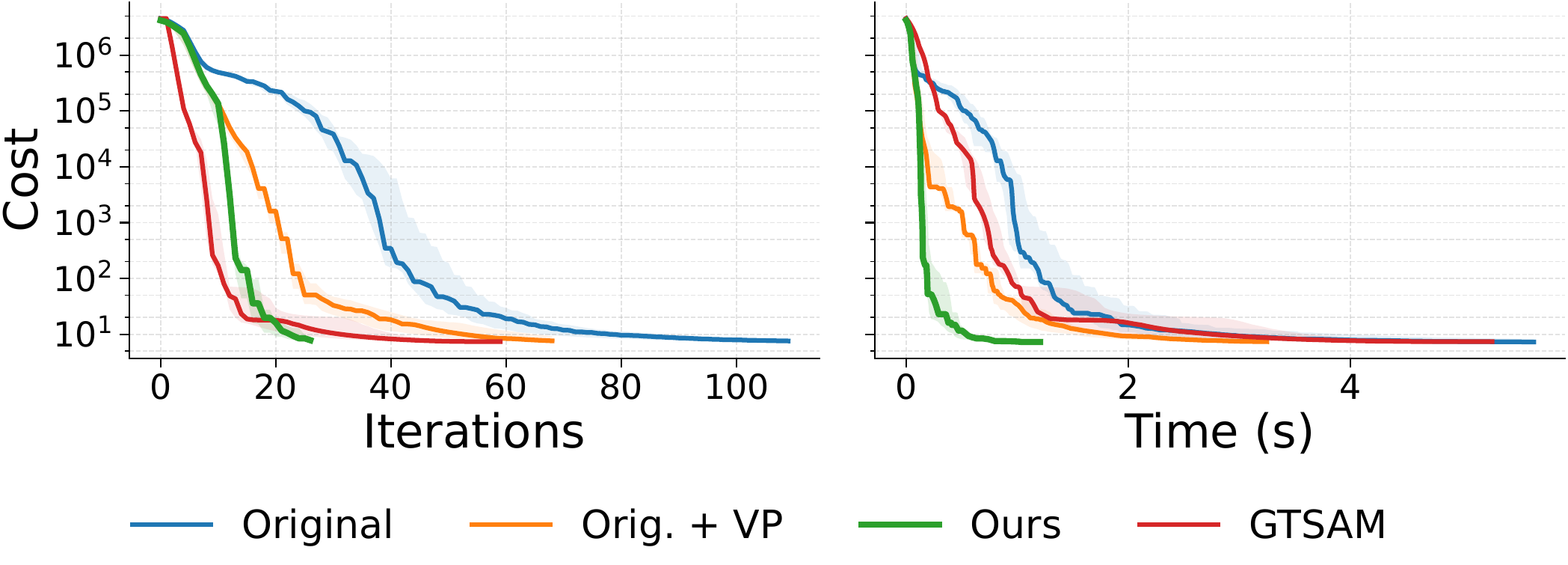}
    }

    \vspace{0.6em}

    \subfloat[SNL (MIT)\label{fig:c}]{
        \includegraphics[width=0.48\linewidth]{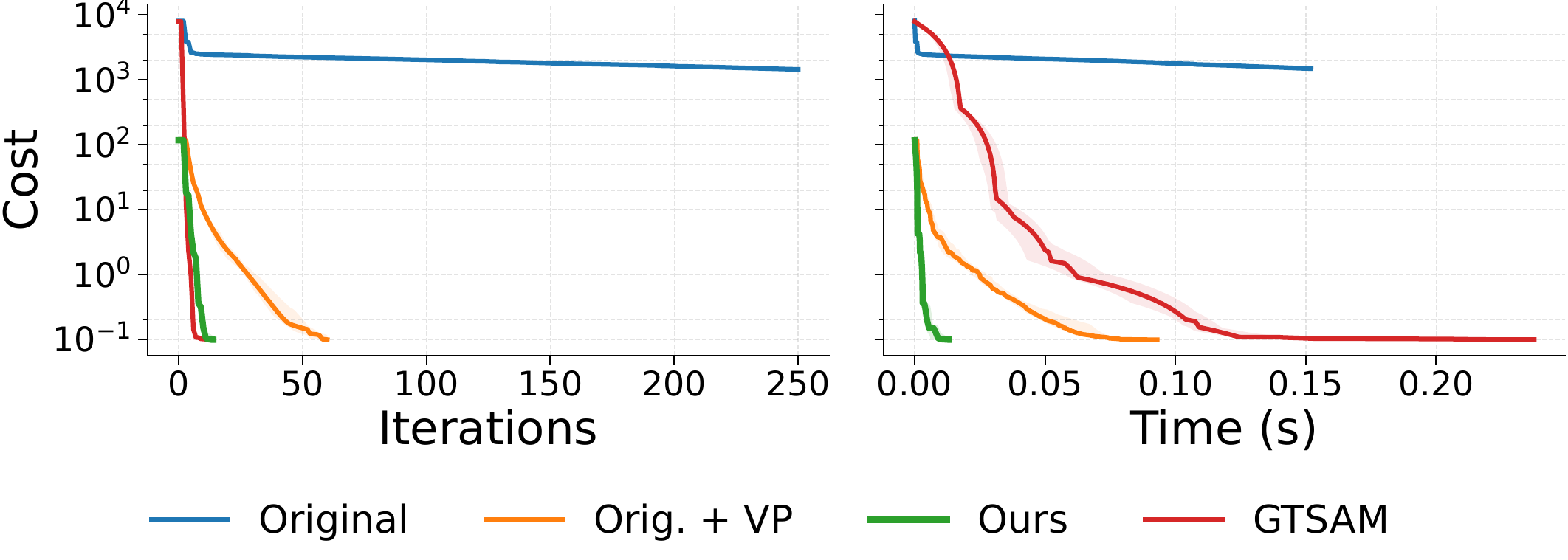}
    }\hfill
    \subfloat[SfM (Mip-NeRF Garden)\label{fig:d}]{
        \includegraphics[width=0.48\linewidth]{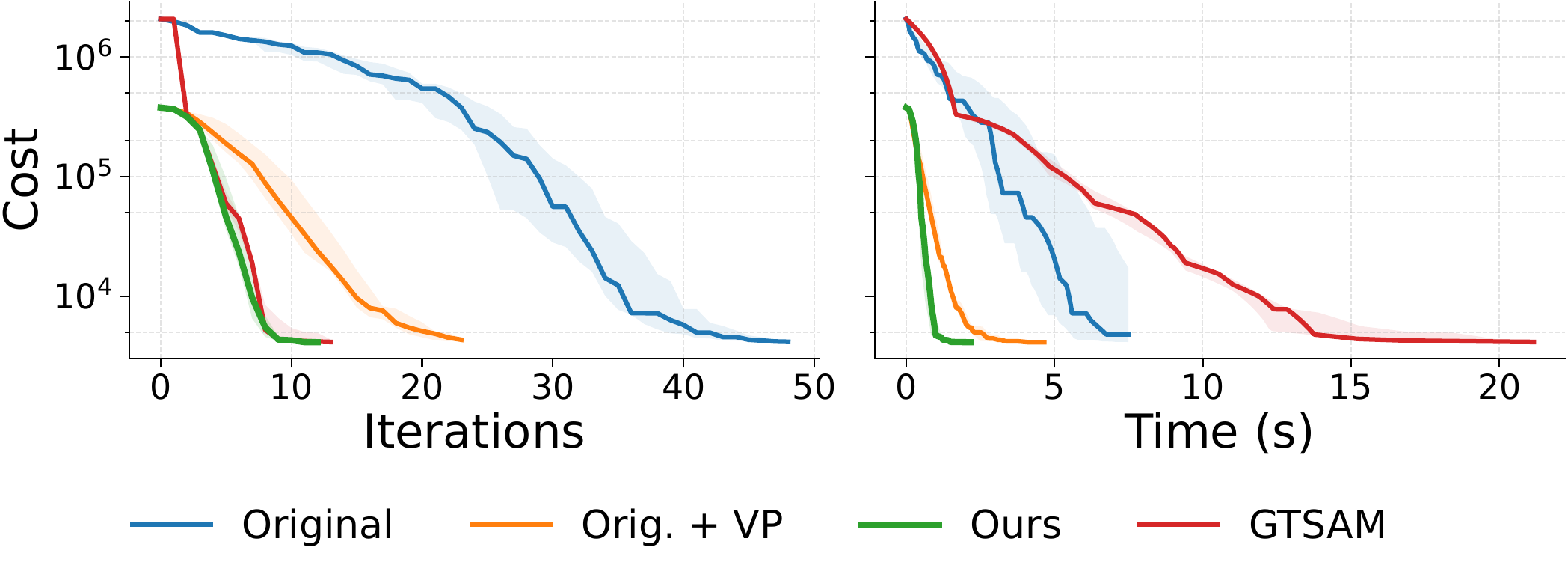}
    }

    \caption{\textbf{Convergence behavior on select problems.}
        Each panel pair shows (left) cost vs.\ iterations and (right) cost vs.\
        time for our method and the baseline methodologies on representative
        datasets chosen from
        (a) pose-graph optimization (Intel),
        (b) RA-SLAM (Single Drone),
        (c) SNL (MIT), and
        (d) SfM (Mip-NeRF Garden). Notably, the most iteration-efficient method
        is not always the most time-efficient, as some methods have higher
        per-iteration costs.
    }
    \label{fig:four-plots}
    \vspace{-1.5em}
\end{figure*}

\subsection{Experiments}

\textbf{Dataset construction.}
The pose graph optimization (PGO) datasets are from \cite{carlone2015Initialization}.
The range-aided SLAM (RA-SLAM) datasets are from \cite{papalia2024Certifiably}.
Sensor network localization (SNL) datasets were generated synthetically from the PGO
datasets by converting all poses to points and all measurements to range 
measurements with identical noise levels. Structure from motion (SfM) datasets 
were generated according to \cite[Sec. IV]{han2025Building}. All dataset names match
original sources.

\textbf{Problem formulations.}
We use problem formulations empirically found to possess benign optimization
landscapes \cite{mcrae2024benign,criscitiello2025sensor}, meaning local
optimization methods can reliably reach global minima from random
initializations. Since all solvers obtain the global minimum, our analysis
avoids complications from local minima.

Specifically, we use formulations from \cite{rosen2019SESync} (PGO),
\cite{papalia2024Certifiably} (RA-SLAM), \cite{halsted22arxiv} (SNL), and
\cite{han2025Building} (SfM). These formulations are closely related:
RA-SLAM generalizes both PGO and SNL. Our SfM formulation differs from 
\cite{han2025Building} by omitting scale estimation in the constraints; the
SfM formulation matches PGO but follows SfM's bipartite sparsity pattern.

\textbf{Experimental setup.}
We generate 5 random initializations per experiment and run each method on all 
5 trials. Solvers run until convergence or 600 seconds elapsed. We compute 
optimal cost by solving a tight convex relaxation and consider a solver 
converged if it reaches within 1\% of this minimum. Times and iteration counts 
in \cref{tab:pgo_wide_gtsam_compare} are medians over 5 trials. Solver failures 
due to non-convergence (not reaching the global optimum within 600 seconds) or 
memory exhaustion are indicated by (\mredx) and (\xmark), respectively.
In all instances where a solver failed in a trial, it was found to
converge in none of the 5 trials.

\subsection{Results}

\textbf{Runtime and iteration efficiency.}
Our approach achieved the lowest runtimes on $39/41$ datasets 
(\cref{tab:pgo_wide_gtsam_compare}), gaining efficiency through both improved 
per-iteration cost and reduced iteration counts (\cref{fig:four-plots}).
On two datasets ({Garage} in PGO and {Plaza 1} in RA-SLAM), GTSAM was faster by 
factors of 1.66 and 1.16, respectively. This is likely because these relatively small
datasets had structure well-suited to GTSAM's direct linear solvers.

Runtime improvements are particularly notable on SfM datasets, which typically 
have far more unconstrained variables (3D points) than constrained variables 
(orientations). Here, our method was often an order of magnitude faster than 
GTSAM and at least 2$\times$ faster than \emph{Original} and \emph{Original + 
VarPro}.


However, despite being faster, our method often requires more iterations than
GTSAM. This likely stems from GTSAM's direct linear solvers, which are robust 
to conditioning and exactly solve subproblems (up to floating point precision), 
enabling more accurate steps per iteration. However, these direct solvers are 
computationally expensive, making GTSAM slower overall despite fewer iterations.

\textbf{Memory efficiency.}
The SfM experiments show our method is more memory efficient than GTSAM, which 
often ran out of memory. This efficiency stems from two properties: (1) our 
reduced problem preserves sparsity and requires no large dense matrices when 
the original problem is sparse (typical case), and (2) we use iterative linear 
solvers (pTCG) that avoid storing large dense matrices, unlike GTSAM's direct 
solvers which require factorizing large matrices.

%% file: tab/tab_experiments.tex
\newcommand{\xmark}{\ensuremath{\text{\ding{55}}}}
\newcommand{\mredx}{\ensuremath{-}}
\newcommand{\goodPct}[1]{
    \textcolor{ForestGreen}{#1}%
}
\newcommand{\badPct}[1]{
    \textcolor{BrickRed}{#1}%
}
\def\best{\bf}
\def\emptyPct{\textemdash}

\newcommand{\dashNA}{\textcolor{gray}{~---\,}}

\newcommand{\speedupCalc}[2]{%
    \begingroup
    \ifstrequal{\detokenize{#1}}{\detokenize{\mredx}}{\dashNA}{%
        \ifstrequal{\detokenize{#1}}{\detokenize{\xmark}}{\dashNA}{%
            \ifstrequal{\detokenize{#2}}{\detokenize{\mredx}}{\dashNA}{%
                \ifstrequal{\detokenize{#2}}{\detokenize{\xmark}}{\dashNA}{%
                    \edef\A{\fpeval{#1}}%
                    \edef\B{\fpeval{#2}}%
                    \ifdim \B pt = 0pt
                        \dashNA
                    \else
                        \edef\ratio{\fpeval{round(\A/\B,2)}}%
                        \ifdim \ratio pt > 1pt
                            \goodPct{\ratio}%
                        \else
                            \badPct{\ratio}%
                        \fi
                    \fi
                }}}}%
    \endgroup
}

\newcommand{\iterCalc}[2]{%
    \begingroup
    \ifstrequal{\detokenize{#1}}{\detokenize{\mredx}}{\dashNA}{%
        \ifstrequal{\detokenize{#1}}{\detokenize{\xmark}}{\dashNA}{%
            \ifstrequal{\detokenize{#2}}{\detokenize{\mredx}}{\dashNA}{%
                \ifstrequal{\detokenize{#2}}{\detokenize{\xmark}}{\dashNA}{%
                    \edef\A{\fpeval{#1}}%
                    \edef\B{\fpeval{#2}}%
                    \ifdim \B pt = 0pt
                        \dashNA
                    \else
                        \edef\ratio{\fpeval{round(\A/\B,2)}}%
                        \ifdim \ratio pt > 1pt
                                {\ratio}%
                        \else
                            \ifdim \ratio pt < 1pt
                                    {\fpeval{round(#1/#2,2)}}%
                            \else
                                \dashNA
                            \fi
                        \fi
                    \fi
                }}}}%
    \endgroup
}

\DeclareRobustCommand{\boldifless}[3]{%
    \begingroup
    \ifstrequal{#1}{\mredx}{#1\endgroup}{%
        \ifstrequal{#1}{\xmark}{#1\endgroup}{%
            \edef\A{\fpeval{#1}}%
            \newif\ifOKB \newif\ifOKC
            \ifstrequal{#2}{\mredx}{\OKBtrue}{%
                \ifstrequal{#2}{\xmark}{\OKBtrue}{%
                    \edef\B{\fpeval{#2}}%
                    \ifdim \A pt > \B pt \OKBfalse \else \OKBtrue \fi
                }}%
            \ifstrequal{#3}{\mredx}{\OKCtrue}{%
                \ifstrequal{#3}{\xmark}{\OKCtrue}{%
                    \edef\C{\fpeval{#3}}%
                    \ifdim \A pt > \C pt \OKCfalse \else \OKCtrue \fi
                }}%
            \ifOKB\ifOKC \textbf{\A}\else \A\fi\else \A\fi
            \endgroup
        }}%
}

\newcommand{\tableRow}[7]{%
    #1
    & \boldifless{#2}{#3}{#4}
    & \boldifless{#3}{#2}{#4}
    & \boldifless{#4}{#2}{#3}
    & #5 & #6 & #7
    & \speedupCalc{#3}{#2}\,/\!\speedupCalc{#4}{#2}
    & \iterCalc{#6}{#5}\,/\,\iterCalc{#7}{#5}
    \\%
}

\def\pgoMultiRow{
    \multirow[c]{8}{*}{\rotatebox[origin=c]{90}{\textbf{PGO}}}
}
\def\raslamMultiRow{
    \multirow[c]{8}{*}{\rotatebox[origin=c]{90}{\textbf{RA-SLAM}}}
}
\def\snlMultiRow{
    \multirow[c]{8}{*}{\rotatebox[origin=c]{90}{\textbf{SNL}}}
}
\def\sfmMultiRow{
    \multirow[c]{17}{*}{\rotatebox[origin=c]{90}{\textbf{SfM}}}
}
\renewcommand{\tableRow}[9]{
    #1
    & \boldifless{#2}{#4}{#5}
    & \boldifless{#3}{#2}{#4}
    & \boldifless{#4}{#2}{#5}
    & \boldifless{#5}{#2}{#4}
    & #6 & #7 & #8 & #9
    & \speedupCalc{#3}{#2}\,/\!\speedupCalc{#4}{#2}\,/\!\speedupCalc{#5}{#2}
    & \iterCalc{#7}{#6}\,/\,\iterCalc{#8}{#6}\,/\,\iterCalc{#9}{#6}
    \\%
}
\begin{table*}[t]
    \vspace{0.5em}
    \centering
    \caption{
        \textbf{Runtime and Iterations Results for all Experiments:}
        Runtime (in seconds) and number of iterations each solver required for
        the proposed method (Ours) and the Original, Original + VarPro, and
        GTSAM approaches across all datasets.
        {The runtime and
        iteration improvement factors are defined as
        $\frac{\text{Baseline}}{\text{Ours}}$.
        Values $>1$ indicate our method required less time/iterations than the
        baseline and $<1$ indicates our method required more (e.g., $2$ means
        our method runs in half the time or with half the iterations).
        Each row's fastest runtime is bolded.  For runtime improvement factors,
        values $>1$ are green and values $<1$ are red.  We use
        (\mredx) to indicate a method failed to converge to the global minimum
        and (\xmark) to indicate a method ran out of memory.}
    }
    \label{tab:pgo_wide_gtsam_compare}
    \resizebox{\linewidth}{!}{%
        \begin{tabular}{ll cccc cccc cc}
            \toprule
                            &                                                                                  & \multicolumn{4}{c}{\textbf{{Runtime (s)}} $\downarrow$} & \multicolumn{4}{c}{\textbf{Solver Iterations} $\downarrow$} & \multicolumn{2}{c}{\textbf{Improvement Factor} $\uparrow$}                                                                                                                                                                                                                                     \\
            \cmidrule(lr){3-6} \cmidrule(lr){7-10} \cmidrule(lr){11-12}
                            & Dataset                                                                          & Ours                                                    & Original                                                    & Orig. + VP                                                 & GTSAM & Ours & Original & Orig. + VP & GTSAM & \makecell[c]{Runtime \\{\scriptsize(Orig.\,/Orig + VP/\,GTSAM)}} & \makecell[c]{Iterations \\{\scriptsize(Orig.\,/Orig + VP/\,GTSAM)}} \\
            \midrule
            \pgoMultiRow    & \tableRow{Intel}{0.19}{0.83}{0.50}{0.63}{21}{69}{40}{12}
                            & \tableRow{Garage}{12.48}{\mredx}{\mredx}{7.51}{219}{\mredx}{\mredx}{63}
                            & \tableRow{Grid3D}{3.60}{10.22}{4.26}{\mredx}{14}{22}{13}{\mredx}
                            & \tableRow{MIT}{0.09}{0.25}{0.16}{0.28}{20}{39}{31}{11}
                            & \tableRow{M3500}{0.72}{11.48}{8.06}{2.85}{31}{183}{120}{22}
                            & \tableRow{City10000}{4.09}{26.09}{10.09}{18.43}{31}{88}{54}{29}
                            & \tableRow{Torus}{0.89}{1.09}{1.13}{37.24}{12}{16}{13}{36}
                            & \tableRow{Sphere}{1.06}{2.06}{1.83}{14.68}{20}{30}{26}{39}
            \midrule
            \raslamMultiRow & \tableRow{TIERS}{10.49}{48.20}{20.96}{\mredx}{67}{133}{74}{\mredx}

                            & \tableRow{Single Drone}{1.03}{5.53}{3.62}{4.91}{32}{130}{79}{57}
                            & \tableRow{Plaza2}{1.45}{5.53}{2.11}{4.73}{37}{85}{34}{57}
                            & \tableRow{Plaza1}{12.76}{22.07}{16.87}{10.91}{62}{140}{74}{63}
                            & \tableRow{MR.CLAM2}{6.17}{\mredx}{13.57}{11.59}{28}{\mredx}{34}{18}
                            & \tableRow{MR.CLAM4}{4.53}{33.09}{11.77}{8.02}{26}{178}{30}{15}
                            & \tableRow{MR.CLAM6}{3.07}{17.03}{5.21}{5.89}{31}{219}{34}{25}
                            & \tableRow{MR.CLAM7}{3.18}{23.42}{7.65}{10.32}{32}{224}{32}{36}

            \midrule
            \snlMultiRow    & \tableRow{Intel}{0.13}{\mredx}{0.78}{\mredx}{32}{\mredx}{121}{\mredx}
                            & \tableRow{Garage}{3.94}{\mredx}{15.29}{\mredx}{245}{\mredx}{249}{\mredx}
                            & \tableRow{Grid3D}{10.56}{\mredx}{16.94}{\mredx}{63}{\mredx}{83}{\mredx}
                            & \tableRow{MIT}{0.01}{\mredx}{0.08}{0.21}{13}{\mredx}{60}{12}
                            & \tableRow{M3500}{0.69}{\mredx}{\mredx}{\mredx}{61}{\mredx}{\mredx}{\mredx}
                            & \tableRow{City10000}{28.46}{\mredx}{\mredx}{\mredx}{208}{\mredx}{\mredx}{\mredx}
                            & \tableRow{Torus}{1.99}{\mredx}{20.83}{\mredx}{40}{\mredx}{137}{\mredx}
                            & \tableRow{Sphere}{2.65}{\mredx}{\mredx}{\mredx}{70}{\mredx}{\mredx}{\mredx}
            \midrule
            \sfmMultiRow
                            & \tableRow{BAL-93}{0.39}{4.50}{1.45}{10.09}{17}{213}{63}{52}
                            & \tableRow{BAL-392}{9.60}{\mredx}{93.32}{\mredx}{21}{\mredx}{149}{\mredx}
                            & \tableRow{BAL-1934}{43.61}{\mredx}{\mredx}{\xmark}{23}{\mredx}{\mredx}{\xmark}
                            & \tableRow{IMC Gate}{18.74}{\mredx}{\mredx}{\xmark}{33}{\mredx}{\mredx}{\xmark}
                            & \tableRow{IMC Temple}{12.59}{91.13}{43.21}{\xmark}{20}{86}{48}{\xmark}
                            & \tableRow{IMC Rome}{55.41}{\mredx}{\mredx}{\xmark}{41}{\mredx}{\mredx}{\xmark}
                            & \tableRow{Rep. Office0-100}{0.33}{2.07}{0.92}{8.39}{11}{40}{20}{10}
                            & \tableRow{Rep. Office1-100}{0.17}{2.32}{0.71}{3.14}{11}{114}{30}{9}
                            & \tableRow{Rep. Room0-100}{0.54}{3.12}{1.40}{12.00}{13}{38}{19}{9}
                            & \tableRow{Rep. Room1-100}{0.46}{4.60}{1.63}{13.63}{12}{62}{37}{13}
                            & \tableRow{Mip-NeRF Garden}{1.48}{9.53}{3.98}{19.76}{11}{48}{25}{12}
                            & \tableRow{Mip-NeRF Room}{17.82}{\mredx}{54.79}{\mredx}{20}{\mredx}{48}{\mredx}
                            & \tableRow{Mip-NeRF Kitchen}{13.03}{\mredx}{83.61}{\mredx}{14}{\mredx}{47}{\mredx}
                            & \tableRow{TUM Room}{15.39}{\mredx}{89.11}{\xmark}{23}{\mredx}{60}{\xmark}
                            & \tableRow{TUM Desk}{8.12}{\mredx}{33.32}{\xmark}{14}{\mredx}{51}{\xmark}
                            & \tableRow{TUM Comp-R}{5.38}{83.09}{20.09}{\xmark}{16}{249}{47}{\xmark}
                            & \tableRow{TUM Comp-T}{10.59}{\mredx}{56.92}{\xmark}{17}{\mredx}{62}{\xmark}

            \bottomrule
        \end{tabular}
    }
    \vspace{-2em}
\end{table*}

%% file: sec/6_conclusion.tex
\section{Conclusion}
\label{sec:conclusion}

This paper introduced a novel method to accelerate the solution of large-scale,
separable nonlinear least-squares problems prevalent in robotic perception.  Our
approach leverages a sparsity-preserving VarPro scheme that can be applied as an
efficient, one-time preprocessing step. The scheme represents the reduced
problem with an efficient matrix-free Schur complement operator that readily
integrates with standard iterative optimization frameworks.

Experiments on diverse problems demonstrated the practical benefits of our
approach. We achieved significant reductions in solver runtime compared to
existing methods, including a prior VarPro technique \cite{khosoussi2016Sparse}.
Our method proved more scalable and memory-efficient than the state-of-the-art
GTSAM solver, most notably on large-scale SfM instances where GTSAM crashed due
to memory constraints.

The core of our method's success lies in its ability to exploit both problem
separability and sparsity, overcoming the challenges posed by gauge symmetries
that have limited previous approaches (see
\cref{sec:prelim:efficient-schur-complement-products} for details).  We also
provided a clear characterization of the class of problems for which our method
is applicable, offering a practical guide for its adoption on new problems.

To support further research and application, our work is released as an
open-source C++ library, including all datasets and baselines from our
experiments.

%% file: bib/references.bib
@string{cvpr  = "Proc. {IEEE} Int. Conf. Computer Vision and Pattern Recognition"}

@string{eccv  = "Eur. Conf. on Computer Vision (ECCV)"}

@string{rss   = "Robotics: Science and Systems (RSS)"}

@string{icra  = "IEEE Intl. Conf. on Robotics and Automation (ICRA)"}

@string{ieee  = "Proc. of the IEEE"}

@misc{weber2024Power,
  title     = {Power Variable Projection for Initialization-Free Large-Scale Bundle Adjustment},
  url       = {http://arxiv.org/abs/2405.05079},
  doi       = {10.48550/arXiv.2405.05079},
  language  = {en},
  urldate   = {2025-08-21},
  publisher = {arXiv},
  author    = {Weber, Simon and Hong, Je Hyeong and Cremers, Daniel},
  month     = aug,
  year      = {2024},
  note      = {arXiv:2405.05079 [cs]}
}

@inproceedings{hong2017Revisiting,
  address   = {Honolulu, HI},
  title     = {Revisiting the Variable Projection Method for Separable Nonlinear Least Squares Problems},
  isbn      = {978-1-5386-0457-1},
  url       = {http://ieeexplore.ieee.org/document/8100112/},
  doi       = {10.1109/CVPR.2017.629},
  language  = {en},
  urldate   = {2025-08-21},
  booktitle = {2017 {IEEE} {Conference} on {Computer} {Vision} and {Pattern} {Recognition} ({CVPR})},
  publisher = {IEEE},
  author    = {Hong, Je Hyeong and Zach, Christopher and Fitzgibbon, Andrew},
  month     = jul,
  year      = {2017},
  pages     = {5939--5947}
}

@incollection{hong2016Projective,
  address   = {Cham},
  title     = {Projective Bundle Adjustment from Arbitrary Initialization Using the Variable Projection Method},
  volume    = {9905},
  isbn      = {978-3-319-46447-3 978-3-319-46448-0},
  url       = {http://link.springer.com/10.1007/978-3-319-46448-0_29},
  language  = {en},
  urldate   = {2025-08-21},
  booktitle = {Computer {Vision} – {ECCV} 2016},
  publisher = {Springer International Publishing},
  author    = {Hong, Je Hyeong and Zach, Christopher and Fitzgibbon, Andrew and Cipolla, Roberto},
  editor    = {Leibe, Bastian and Matas, Jiri and Sebe, Nicu and Welling, Max},
  year      = {2016},
  doi       = {10.1007/978-3-319-46448-0_29},
  note      = {Series Title: Lecture Notes in Computer Science},
  pages     = {477--493}
}

@article{golub2003Separable,
  title      = {Separable nonlinear least squares: the variable projection method and its applications},
  volume     = {19},
  issn       = {0266-5611, 1361-6420},
  shorttitle = {Separable nonlinear least squares},
  url        = {https://iopscience.iop.org/article/10.1088/0266-5611/19/2/201},
  doi        = {10.1088/0266-5611/19/2/201},
  language   = {en},
  number     = {2},
  urldate    = {2025-08-21},
  journal    = {Inverse Problems},
  author     = {Golub, Gene and Pereyra, Victor},
  month      = apr,
  year       = {2003},
  pages      = {R1--R26}
}

@article{matthews2004Active,
  title     = {Active Appearance Models Revisited},
  volume    = {60},
  copyright = {https://www.springernature.com/gp/researchers/text-and-data-mining},
  issn      = {0920-5691, 1573-1405},
  url       = {https://link.springer.com/10.1023/B:VISI.0000029666.37597.d3},
  doi       = {10.1023/B:VISI.0000029666.37597.d3},
  language  = {en},
  number    = {2},
  urldate   = {2025-08-22},
  journal   = {International Journal of Computer Vision},
  author    = {Matthews, Iain and Baker, Simon},
  month     = nov,
  year      = {2004},
  pages     = {135--164}
}

@inproceedings{carlone2015Initialization,
  address    = {Seattle, WA, USA},
  title      = {Initialization techniques for {3D} {SLAM}: {A} survey on rotation estimation and its use in pose graph optimization},
  isbn       = {978-1-4799-6923-4},
  shorttitle = {Initialization techniques for {3D} {SLAM}},
  url        = {http://ieeexplore.ieee.org/document/7139836/},
  doi        = {10.1109/ICRA.2015.7139836},
  language   = {en},
  urldate    = {2025-08-22},
  booktitle  = {2015 {IEEE} {International} {Conference} on {Robotics} and {Automation} ({ICRA})},
  publisher  = {IEEE},
  author     = {Carlone, Luca and Tron, Roberto and Daniilidis, Kostas and Dellaert, Frank},
  month      = may,
  year       = {2015},
  pages      = {4597--4604}
}

@inproceedings{kummerle2011G2o,
  address    = {Shanghai, China},
  title      = {G$^{\textrm{2}}$o: {A} general framework for graph optimization},
  isbn       = {978-1-61284-386-5},
  shorttitle = {G$^{\textrm{2}}$o},
  url        = {http://ieeexplore.ieee.org/document/5979949/},
  doi        = {10.1109/ICRA.2011.5979949},
  language   = {en},
  urldate    = {2025-08-22},
  booktitle  = {2011 {IEEE} {International} {Conference} on {Robotics} and {Automation}},
  publisher  = {IEEE},
  author     = {Kummerle, Rainer and Grisetti, Giorgio and Strasdat, Hauke and Konolige, Kurt and Burgard, Wolfram},
  month      = may,
  year       = {2011},
  pages      = {3607--3613}
}

@inproceedings{okatani2011Efficient,
  address   = {Barcelona, Spain},
  title     = {Efficient algorithm for low-rank matrix factorization with missing components and performance comparison of latest algorithms},
  isbn      = {978-1-4577-1102-2 978-1-4577-1101-5 978-1-4577-1100-8},
  url       = {http://ieeexplore.ieee.org/document/6126324/},
  doi       = {10.1109/ICCV.2011.6126324},
  language  = {en},
  urldate   = {2025-08-22},
  booktitle = {2011 {International} {Conference} on {Computer} {Vision}},
  publisher = {IEEE},
  author    = {Okatani, Takayuki and Yoshida, Takahiro and Deguchi, Koichiro},
  month     = nov,
  year      = {2011},
  pages     = {842--849}
}

@inproceedings{woodford2020Large,
  title     = {Large Scale Photometric Bundle Adjustment},
  url       = {http://arxiv.org/abs/2008.11762},
  doi       = {10.48550/arXiv.2008.11762},
  language  = {en},
  booktitle = {2020 {British} {Machine} {Vision} {Conference}},
  urldate   = {2025-09-06},
  publisher = {arXiv},
  author    = {Woodford, Oliver J. and Rosten, Edward},
  month     = sep,
  year      = {2020},
  note      = {arXiv:2008.11762 [cs]}
}

@article{khosoussi2016Sparse,
  title = {A Sparse Separable {SLAM} Back-End},
  volume    = {32},
  copyright = {https://ieeexplore.ieee.org/Xplorehelp/downloads/license-information/IEEE.html},
  issn      = {1552-3098, 1941-0468},
  url       = {http://ieeexplore.ieee.org/document/7592861/},
  doi       = {10.1109/TRO.2016.2609394},
  language  = {en},
  number    = {6},
  urldate   = {2025-08-22},
  journal   = {IEEE Transactions on Robotics},
  author    = {Khosoussi, Kasra and Huang, Shoudong and Dissanayake, Gamini},
  month     = dec,
  year      = {2016},
  pages     = {1536--1549}
}

@incollection{agarwal2010Bundle,
  address   = {Berlin, Heidelberg},
  title     = {Bundle Adjustment in the Large},
  volume    = {6312},
  copyright = {http://www.springer.com/tdm},
  isbn      = {978-3-642-15551-2 978-3-642-15552-9},
  url       = {http://link.springer.com/10.1007/978-3-642-15552-9_3},
  language  = {en},
  urldate   = {2025-09-06},
  booktitle = {Computer {Vision} – {ECCV} 2010},
  publisher = {Springer Berlin Heidelberg},
  author    = {Agarwal, Sameer and Snavely, Noah and Seitz, Steven M. and Szeliski, Richard},
  year      = {2010},
  doi       = {10.1007/978-3-642-15552-9_3},
  note      = {Series Title: Lecture Notes in Computer Science},
  pages     = {29--42}
}

@software{agarwal2022ceres,
  author  = {Agarwal, Sameer and Mierle, Keir and The Ceres Solver Team},
  title   = {{Ceres Solver}},
  license = {Apache-2.0},
  url     = {https://github.com/ceres-solver/ceres-solver},
  version = {2.2},
  year    = {2023},
  month   = {10}
}

@article{dellaert2012factor,
  title   = {Factor graphs and {GTSAM}: A hands-on introduction},
  author  = {Dellaert, Frank},
  journal = {Georgia Institute of Technology, Tech. Rep},
  volume  = {2},
  number  = {4},
  year    = {2012}
}

@article{rosen2019SESync,
  title      = {{SE}-{Sync}: {A} certifiably correct algorithm for synchronization over the special {Euclidean} group},
  volume     = {38},
  issn       = {0278-3649, 1741-3176},
  shorttitle = {{SE}-{Sync}},
  url        = {https://journals.sagepub.com/doi/10.1177/0278364918784361},
  doi        = {10.1177/0278364918784361},
  language   = {en},
  number     = {2-3},
  urldate    = {2025-08-29},
  journal    = {The International Journal of Robotics Research},
  author     = {Rosen, David M and Carlone, Luca and Bandeira, Afonso S and Leonard, John J},
  month      = mar,
  year       = {2019},
  pages      = {95--125}
}

@article{papalia2024Certifiably,
  title     = {Certifiably Correct Range-Aided {SLAM}},
  volume    = {40},
  copyright = {https://ieeexplore.ieee.org/Xplorehelp/downloads/license-information/IEEE.html},
  issn      = {1552-3098, 1941-0468},
  url       = {https://ieeexplore.ieee.org/document/10665918/},
  doi       = {10.1109/TRO.2024.3454430},
  language  = {en},
  urldate   = {2025-08-29},
  journal   = {IEEE Transactions on Robotics},
  author    = {Papalia, Alan and Fishberg, Andrew and O'Neill, Brendan W. and How, Jonathan P. and Rosen, David M. and Leonard, John J.},
  year      = {2024},
  pages     = {4265--4283}
}

@article{halsted22arxiv,
  author  = {Halsted, Trevor and Schwager, Mac},
  title   = {The {Riemannian} Elevator for certifiable distance-based localization},
  year    = {2022},
  journal = {Preprint},
  note    = {{Accessed:} Jan. 20, 2023. [Online] Available \url{https://msl.stanford.edu/papers/halsted_riemannian_2022.pdf}}
}

@article{han2025Building,
  title     = {Building {Rome} with Convex Optimization},
  url       = {http://arxiv.org/abs/2502.04640},
  doi       = {10.48550/arXiv.2502.04640},
  urldate   = {2025-08-29},
  publisher = {arXiv},
  author    = {Han, Haoyu and Yang, Heng},
  journal   = rss,
  month     = jul,
  year      = {2025},
  note      = {arXiv:2502.04640 [cs]}
}

@article{strang2024Elimination,
  title    = {Elimination and Factorization},
  volume   = {97},
  issn     = {0025-570X, 1930-0980},
  url      = {https://www.tandfonline.com/doi/full/10.1080/0025570X.2024.2401295},
  doi      = {10.1080/0025570X.2024.2401295},
  language = {en},
  number   = {5},
  urldate  = {2025-09-02},
  journal  = {Mathematics Magazine},
  author   = {Strang, Gilbert},
  month    = oct,
  year     = {2024},
  pages    = {484--487}
}

@article{strang2022Three,
  title    = {Three matrix factorizations from the steps of elimination},
  volume   = {20},
  issn     = {0219-5305, 1793-6861},
  url      = {https://www.worldscientific.com/doi/10.1142/S0219530522400061},
  doi      = {10.1142/S0219530522400061},
  language = {en},
  number   = {06},
  urldate  = {2025-09-02},
  journal  = {Analysis and Applications},
  author   = {Strang, Gilbert and Drucker, Daniel},
  month    = nov,
  year     = {2022},
  pages    = {1147--1157}
}

@book{zhang2006schur,
  title     = {The {Schur} complement and its applications},
  author    = {Zhang, Fuzhen},
  volume    = {4},
  year      = {2006},
  publisher = {Springer Science \& Business Media}
}

@article{ruhe1980algorithms,
  title     = {Algorithms for separable nonlinear least squares problems},
  author    = {Ruhe, Axel and Wedin, Per {\AA}ke},
  journal   = {SIAM review},
  volume    = {22},
  number    = {3},
  pages     = {318--337},
  year      = {1980},
  publisher = {SIAM}
}

@book{saad2003iterative,
  title     = {Iterative methods for sparse linear systems},
  author    = {Saad, Yousef},
  year      = {2003},
  publisher = {SIAM}
}

@article{tian2020asynchronous,
  title     = {Asynchronous and parallel distributed pose graph optimization},
  author    = {Tian, Yulun and Koppel, Alec and Bedi, Amrit Singh and How, Jonathan P},
  journal   = {IEEE Robotics and Automation Letters},
  volume    = {5},
  number    = {4},
  pages     = {5819--5826},
  year      = {2020},
  publisher = {IEEE}
}

@inproceedings{mcgann2024asynchronous,
  title        = {Asynchronous distributed smoothing and mapping via on-manifold consensus {ADMM}},
  author       = {McGann, Daniel and Lassak, Kyle and Kaess, Michael},
  booktitle    = {2024 IEEE International Conference on Robotics and Automation (ICRA)},
  pages        = {4577--4583},
  year         = {2024},
  organization = {IEEE}
}

@article{fan2023majorization,
  title     = {Majorization minimization methods for distributed pose graph optimization},
  author    = {Fan, Taosha and Murphey, Todd D.},
  journal   = {IEEE Transactions on Robotics},
  volume    = {40},
  pages     = {22--42},
  year      = {2024},
  publisher = {IEEE},
  doi       = {10.1109/TRO.2023.3324818}
}

@inproceedings{papalia2023score,
  author    = {Papalia, Alan and Morales, Joseph and Doherty, Kevin J. and Rosen, David M. and Leonard, John J.},
  booktitle = {2023 IEEE International Conference on Robotics and Automation (ICRA)},
  title     = {{SCORE}: A Second-Order Conic Initialization for Range-Aided {SLAM}},
  year      = {2023},
  volume    = {},
  number    = {},
  pages     = {10637-10644},
  doi       = {10.1109/ICRA48891.2023.10160787}
}

@incollection{triggs2000Bundle,
  address   = {Berlin, Heidelberg},
  title     = {Bundle Adjustment—A Modern Synthesis},
  volume    = {1883},
  isbn      = {978-3-540-67973-8 978-3-540-44480-0},
  url       = {https://link.springer.com/10.1007/3-540-44480-7_21},
  language  = {en},
  urldate   = {2025-09-06},
  booktitle = {Vision {Algorithms}: {Theory} and {Practice}},
  publisher = {Springer Berlin Heidelberg},
  author    = {Triggs, Bill and McLauchlan, Philip F. and Hartley, Richard I. and Fitzgibbon, Andrew W.},
  editor    = {Goos, Gerhard and Hartmanis, Juris and Van Leeuwen, Jan and Triggs, Bill and Zisserman, Andrew and Szeliski, Richard},
  year      = {2000},
  doi       = {10.1007/3-540-44480-7_21},
  note      = {Series Title: Lecture Notes in Computer Science},
  pages     = {298--372}
}

@article{dellaert2017factor,
  title     = {Factor graphs for robot perception},
  author    = {Dellaert, Frank and Kaess, Michael and others},
  journal   = {Foundations and Trends{\textregistered} in Robotics},
  volume    = {6},
  number    = {1-2},
  pages     = {1--139},
  year      = {2017},
  publisher = {Now Publishers, Inc.}
}

@article{ebadi2023present,
  title     = {Present and future of {SLAM} in extreme environments: The {DARPA SUBT} challenge},
  author    = {Ebadi, Kamak and Bernreiter, Lukas and Biggie, Harel and Catt, Gavin and Chang, Yun and Chatterjee, Arghya and Denniston, Christopher E and Desch{\^e}nes, Simon-Pierre and Harlow, Kyle and Khattak, Shehryar and others},
  journal   = {IEEE Transactions on Robotics},
  volume    = {40},
  pages     = {936--959},
  year      = {2023},
  publisher = {IEEE}
}

@article{kunze2018artificial,
  title     = {Artificial intelligence for long-term robot autonomy: A survey},
  author    = {Kunze, Lars and Hawes, Nick and Duckett, Tom and Hanheide, Marc and Krajn{\'\i}k, Tom{\'a}{\v{s}}},
  journal   = {IEEE Robotics and Automation Letters},
  volume    = {3},
  number    = {4},
  pages     = {4023--4030},
  year      = {2018},
  publisher = {IEEE}
}

@article{tranzatto2022cerberus,
  title     = {Cerberus in the {DARPA} subterranean challenge},
  author    = {Tranzatto, Marco and Miki, Takahiro and Dharmadhikari, Mihir and Bernreiter, Lukas and Kulkarni, Mihir and Mascarich, Frank and Andersson, Olov and Khattak, Shehryar and Hutter, Marco and Siegwart, Roland and others},
  journal   = {Science Robotics},
  volume    = {7},
  number    = {66},
  pages     = {eabp9742},
  year      = {2022},
  publisher = {American Association for the Advancement of Science}
}

@article{cadena2017past,
  title     = {Past, present, and future of simultaneous localization and mapping: Toward the robust-perception age},
  author    = {Cadena, Cesar and Carlone, Luca and Carrillo, Henry and Latif, Yasir and Scaramuzza, Davide and Neira, Jos{\'e} and Reid, Ian and Leonard, John J},
  journal   = {IEEE Transactions on robotics},
  volume    = {32},
  number    = {6},
  pages     = {1309--1332},
  year      = {2017},
  publisher = {IEEE}
}

@inproceedings{schonberger2016structure,
  title     = {Structure-from-motion revisited},
  author    = {Schonberger, Johannes L. and Frahm, Jan-Michael},
  booktitle = {Proceedings of the IEEE conference on computer vision and pattern recognition},
  pages     = {4104--4113},
  year      = {2016}
}

@inproceedings{certi_fgo,
  title     = {Simplifying Certifiable Estimation: A Factor Graph Optimization Approach},
  author    = {Xu, Zhexin and Sanderson, Nikolas R. and Rosen, David M.},
  booktitle = icra,
  year      = {2025},
  note      = {{W}orkshop: Robotics in the Wild},
  url       = {https://dartmouthrobotics.github.io/icra-2025-robots-wild/spotlight-papers/icra-2025-robots-wild-16.pdf}
}

@article{mao2007wireless,
  title     = {Wireless sensor network localization techniques},
  author    = {Mao, Guoqiang and Fidan, Bar{\i}{\c{s}} and Anderson, Brian DO},
  journal   = {Computer networks},
  volume    = {51},
  number    = {10},
  pages     = {2529--2553},
  year      = {2007},
  publisher = {Elsevier}
}

@book{golub2013matrix,
  title     = {Matrix computations},
  author    = {Golub, Gene H. and Van Loan, Charles F.},
  year      = {2013},
  publisher = {JHU press}
}

@article{absil2007trust,
  title     = {Trust-region methods on {Riemannian} manifolds},
  author    = {Absil, Pierre-Antoine and Baker, Christopher G.. and Gallivan, Kyle A.},
  journal   = {Foundations of Computational Mathematics},
  volume    = {7},
  number    = {3},
  pages     = {303--330},
  year      = {2007},
  publisher = {Springer}
}

@book{chung1997spectral,
  title     = {Spectral graph theory},
  author    = {Chung, Fan RK},
  volume    = {92},
  year      = {1997},
  publisher = {American Mathematical Soc.}
}

@book{nocedal2006numerical,
  title     = {Numerical optimization},
  author    = {Nocedal, Jorge and Wright, Stephen J.},
  year      = {2006},
  publisher = {Springer}
}

@article{barham1972algorithm,
  title     = {An algorithm for least squares estimation of nonlinear parameters when some of the parameters are linear},
  author    = {Barham, Richard H. and Drane, Wanzer},
  journal   = {Technometrics},
  volume    = {14},
  number    = {3},
  pages     = {757--766},
  year      = {1972},
  publisher = {Taylor \& Francis}
}

@article{luo2010semidefinite,
  title     = {Semidefinite relaxation of quadratic optimization problems},
  author    = {Luo, Zhi-Quan and Ma, Wing-Kin and So, Anthony Man-Cho and Ye, Yinyu and Zhang, Shuzhong},
  journal   = {IEEE Signal Processing Magazine},
  volume    = {27},
  number    = {3},
  pages     = {20--34},
  year      = {2010},
  publisher = {IEEE}
}

@article{coleman1987null,
  title     = {The null space problem {II}. Algorithms},
  author    = {Coleman, Thomas F. and Pothen, Alex},
  journal   = {{SIAM} Journal on Algebraic Discrete Methods},
  volume    = {8},
  number    = {4},
  pages     = {544--563},
  year      = {1987},
  publisher = {SIAM}
}

@book{boumal2023introduction,
  title     = {An introduction to optimization on smooth manifolds},
  author    = {Boumal, Nicolas},
  year      = {2023},
  publisher = {Cambridge University Press}
}

@article{criscitiello2025sensor,
  title={Sensor network localization has a benign landscape after low-dimensional relaxation},
  author={Criscitiello, Christopher and McRae, Andrew D. and Rebjock, Quentin and Boumal, Nicolas},
  journal={arXiv preprint arXiv:2507.15662},
  year={2025}
}

@article{mcrae2024benign,
  title={Benign landscapes of low-dimensional relaxations for orthogonal synchronization on general graphs},
  author={McRae, Andrew D. and Boumal, Nicolas},
  journal={SIAM Journal on Optimization},
  volume={34},
  number={2},
  pages={1427--1454},
  year={2024},
  publisher={SIAM}
}
